\documentclass{article}
\usepackage[utf8]{inputenc}
\usepackage[margin=1in]{geometry}
\usepackage{setspace}
\usepackage{graphicx}
\usepackage{natbib}
\usepackage{hyperref}
\usepackage{amsmath}
\usepackage{booktabs}
\hypersetup{colorlinks,allcolors=black}

\title{Robustness is Important: Limitations of LLMs for Data Fitting}
\author{Hejia Liu$^1$, Mochen Yang$^1$, Gediminas Adomavicius$^1$ \\
$^1$ Carlson School of Management, University of Minnesota}
\date{Current Draft: 10/27/2025}

\begin{document}
%%%%%%%%%%%%%%%%
\doublespacing
\maketitle

\begin{abstract}
    Large Language Models (LLMs) are being applied in a wide array of settings, well beyond typical language-oriented use cases. In particular, LLMs are increasingly used as a plug-and-play method for fitting data and generating predictions. Prior work has shown that LLMs, via in-context learning or supervised fine-tuning, can perform competitively with many tabular supervised learning techniques on predictive performance. However, we identify a critical vulnerability of using LLMs for data fitting -- making changes to data representation that are completely irrelevant to the underlying learning task can drastically alter LLMs' predictions on the same data. For example, simply changing variable names can sway the size of prediction error by as much as 82\% in certain settings. Such prediction sensitivity with respect to task-irrelevant variations manifests under both in-context learning and supervised fine-tuning, for both close-weight and open-weight general-purpose LLMs. Moreover, by examining the attention scores of an open-weight LLM, we discover a non-uniform attention pattern: training examples and variable names/values occupying certain positions in the prompt receive more attention when generating output tokens, even though fundamentally there should not be different emphasis a priori on specific data rows / columns. This partially explains the sensitivity due to task-irrelevant variations. We also consider a state-of-the-art tabular foundation model (TabPFN) trained specifically for data fitting. Despite being explicitly designed to achieve prediction robustness, TabPFN is still not immune to task-irrelevant variations. Overall, despite LLMs' impressive predictive capabilities, currently they lack even the basic level of robustness to be used as a principled data-fitting tool.
\end{abstract}

\section{Introduction}
Large language models (LLMs), as one of the most representative Generative AI applications, have been adopted across a wide variety of domains and tasks. While the majority of LLMs' use cases are expectedly language-related (translation, Q\&A, writing, ideation, etc.), there are also increasing uses of LLMs for tasks not apparently associated with language. We focus on one such practice, where LLM is used as an ``off-the-shelf" \textit{data-fitting tool} to generate predictions of interest. Data fitting refers to the task of using sets $\{ (\boldsymbol{X},Y)\}$ of numerical input values $\boldsymbol{X}$ and target values $Y$ to learn (fit) a model $f(.)$ -- i.e., $Y \sim f(\boldsymbol{X})$ -- that can generate accurate predictions given inputs. For example, LLMs have been used for predicting property prices based on property features (e.g., number of bedrooms,  bathrooms, etc.) and location information \cite{tanlamai2024generative}; performing time series forecasting and numeric prediction tasks in general by treating input data as a sequence of string tokens \cite{gruver2023large,jin2023time,das2024decoder,goswami2024moment,vacareanu2024words}; detecting anomalies in time series data \cite{alnegheimish2024large,dong2024can,liu2025large}; generating synthetic data in data-hungry domains \cite{seedat2023curated,wang2024large, brand2023using, isomura2024llmovertab}; supporting causal effect estimation by predicting counterfactual outcomes \citep{huynhimproving}; and predicting operational metrics in complex systems \citep{akhauri2025performance}. Essentially, LLMs act as a drop-in replacement for traditional, tabular supervised learning techniques to generate predictions based on input data (See \textit{SI Appendix Table S1} for more details about these examples). Importantly, in all of these examples, LLMs are \textit{not} being used as a coding agent to simply generate and/or execute code that applies tabular supervised learning techniques; instead, \textit{LLMs act as the data-fitting tool itself}.

A priori, it is unclear whether LLMs should be used for data fitting and predictions. On one hand, hailed as a general-purpose technology \cite{eloundou2024gpt}, there is evidence that LLMs' impressive capabilities extend beyond language tasks and can perform data fitting quite well. Prior work has shown that LLMs, with just a few input-output examples included in the prompt (i.e., using few-shot prompting), can outperform many tabular supervised learning techniques for regression tasks \citep{vacareanu2024words}. Furthermore, LLMs can select features based on semantic meaning of feature names, which may further benefit data fitting \citep{jeong2024llm}. On the other hand, LLMs vs. tabular supervised learning techniques perform ``data fitting" and ``prediction" in fundamentally different ways. Tabular supervised learning techniques perform data fitting by following well-defined algorithmic procedures that fit parametric functions (e.g., linear regression) or non-parametric functions (e.g., random forest) to training data. The trained models then produce predictions by directly applying the learned function (i.e., plug in the feature values of new input  data). In contrast, the vast majority of LLMs today are based on the transformer architecture, which auto-regressively generates the next token based on prior tokens. With this architecture, LLMs perform data fitting by (i) having training examples in the prompt, namely in-context learning \cite{brown2020language}, or by (ii) supervised fine-tuning using the training data \cite{hu2021lora}.\footnote{We do not consider the use case where LLMs {\em generate and/or execute code} to build predictive models because, as discussed before, we observe an increasingly prevalent use of \textit{LLMs acting as a data-fitting tool itself}, and that is the focus of our study.} They generate predictions one token at a time, where each token may correspond to only a textual segment of the prediction value. As an illustrative example, consider how an LLM generates a numerical prediction value. To generate a prediction value of 0.1875, OpenAI's GPT-4o model needs to generate 4 tokens precisely in the following order: ``0", ``.", ``187", and ``5".\footnote{This example can be replicated at: \url{https://platform.openai.com/tokenizer}} These fundamental differences raise the question whether LLMs provide a principled or robust way to fit numerical data and obtain predictions.

In this work, we offer a critical assessment of using LLMs for data fitting. We construct a numeric prediction task and create synthetic data from scratch. Instead of using publicly available datasets, which LLMs may have been exposed to during training, using synthetic data enables a clean evaluation of LLM's data-fitting capabilities. We experiment with general-purpose LLMs that are both close-weight (GPT-4o-mini) and open-weight (Llama-3-8B-instruct) as well as special-purpose tabular foundation models \citep[TabPFN,][]{hollmann2025accurate}. Consistent with prior work \cite{gruver2023large,vacareanu2024words}, we find that LLMs perform reasonably well (especially after supervised fine-tuning), achieving performance that is competitive with commonly-used tabular supervised learning techniques.

However, just because LLMs \textit{can} be used for data fitting does not mean they \textit{should} be. A principled data-fitting technique not only needs to achieve reasonable predictive performance, but also needs to possess at least a basic level of \textit{robustness} so as not to be affected by factors that are fundamentally unrelated to the task itself. Our analyses show that LLMs have alarmingly poor prediction robustness. In particular, we apply several different types of {\em task-irrelevant} variations -- such as altering variable names, variable (column) order, example (row) order, and data format -- none of which technically should have any effect on the underlying data-fitting task. Canonical tabular supervised learning techniques are, essentially by their design, either completely unaffected or affected to a minor degree explained by randomness (e.g., due to different random seeds that control certain stochastic aspects of the learning procedure). In contrast, LLMs' predictions for the same data can change dramatically, causing the size of prediction error to sway as much as 82\% in some settings. Further statistical tests confirm that such prediction sensitivity is indeed attributable to the task-irrelevant variations and cannot be explained by randomness in the LLM token generation process.

We also investigate the attention scores in an open-weight LLM to explore the potential sources of LLMs' poor prediction robustness. By summing up the attention scores from output tokens (which make up the predicted value) to each of the in-context training examples, we discover a phenomenon of ``U-shaped" attention distribution, where the first and last training examples receive more attention than examples in the middle. Moreover, this phenomenon also manifests within in-context examples -- variable names and values located towards the beginning or end of a given example receive more attention than other positions. Such uneven attention distributions allow training examples and variables that happen to occupy the beginning / end positions in a prompt to have unduly large influence, thereby contributing to substantial variations in LLMs' predictions.

Besides general-purpose LLMs, there are growing efforts to build specialized foundation models for tabular data \cite[``tabular LLM" or ``tabular foundation models",][]{sui2024table,hollmann2025accurate}. We repeat our evaluations on a state-of-the-art tabular foundation model, TabPFN \cite{hollmann2025accurate}, which adopts a modified attention mechanism as an attempt to facilitate the invariance of predictions with respect to row order and variable order. Nonetheless, we find that it is still not immune to task-irrelevant variations. 

Our results reveal a fundamental challenge of using LLMs for data fitting: prediction robustness with respect to task-irrelevant variations, which is generally taken for granted when using tabular supervised learning techniques, can hardly be guaranteed in LLMs; at best, it can only be partially engineered in special-purpose foundation models. As an analogy, using LLMs for data fitting currently can be viewed like using a calculator for adding numbers, where the calculator produces a substantially different sum when given the same exact set of numbers but in a different order. Thus, we caution against the increasingly prevalent use of LLMs as a black-box data-fitting tool -- prediction quality cannot be ensured due to lack of robustness with respect to task-irrelevant variations.

It is crucial to note that the focus of our study is \textit{not} on improving LLMs' data-fitting and prediction capabilities per se. If one takes a very narrow focus on predictive performance and blissfully ignores anything else, one could potentially lean on the increasingly powerful tabular foundation models, or even an ``ensemble" of LLMs (e.g., generate multiple predictions with varying LLMs/prompts and then take an average). However, we maintain the position that data-fitting is fundamentally about ``learning from data", from which good predictive performance emerges as a result \citep{vapnik2013nature}. As such, we use the data-fitting task as an important lens through which we seek \textit{\textbf{to better understand the limitations of current LLMs as a paradigm of learning}}.
%Beyond data fitting, it is worth highlighting that our findings also have broader implications for current LLMs as a paradigm of learning. We take the data-fitting task not as the ultimate objective per se but as an opportunity to better understand the limitations of LLMs. 
The ability to disentangle what is relevant vs. irrelevant for a task is widely recognized as a fundamental requirement for forming abstraction \citep{giere2010explaining,weisberg2012simulation}, learning conceptual categories \citep{rosch2024principles}, cognitive processing \citep[e.g., the selective attention theory][]{broadbent1958perception,johnston1986selective}, all of which are important in problem solving. When LLMs change their predictions in response to task-irrelevant variations, it raises the question whether LLMs possess basic ``competence'' for learning from data. Together with recent work showing that inconsequential changes in prompting can significantly affect LLM performance in solving math problems \citep{mirzadeh2024gsm} and imitating human behaviors \citep{gao2025take}, we highlight that the use of current LLMs as principled learning and problem-solving tools requires careful reconsideration.

\section{Materials and Methods}

Because modern LLMs have been trained with massive datasets collected from the Internet, it is inappropriate to evaluate LLMs' data-fitting capabilities on publicly available datasets which may have already been exposed to the LLMs during training. Therefore, we create synthetic data from scratch for all experiments in this paper. Specifically, we simulate the following linear data generation process for $p$ input variables:
\begin{equation}
\label{eq:DPG}
Y = \sum_{i=1}^p \beta_i X_i + \varepsilon
\end{equation}
In the baseline setting, $p=10$ and each input variable $X_i$ is randomly and independently sampled from either $N(0,1)$ or $\exp(1)$; coefficients $\{\beta_1, \ldots, \beta_p\}$ are sampled from $Unif[0,1]$ and fixed; and $\varepsilon$ is randomly sampled from $N(0,0.1^2)$. We choose such a simple linear data-generation process for two reasons. First, its parametric form gives us fine-grained control over different aspects of the data-fitting task, such as number of input variables ($p$) and distributional characteristics of these variables. Second, it represents an ``easy" data-fitting task. If LLMs exhibit problematic data-fitting capabilities under this setup, it is more likely due to a fundamental limitation of current LLMs rather than difficulty of the data-fitting task. We use this data-generation process to simulate a training dataset of 4,000 instances and a testing dataset of 1,000 instances.

We focus on two specific data-fitting capabilities. We first evaluate the \textit{performance level}, i.e., how well LLMs fit data and generate predictions compared to tabular supervised learning techniques. This determines whether LLMs have the capability to fit data accurately. More importantly, we also evaluate \textit{task-irrelevant prediction sensitivity}, i.e., whether LLM-generated predictions are robust with respect to changes in the inputs that are irrelevant to the data-fitting tasks (and are extraneous for tabular supervised learning techniques). This sheds light on whether LLMs \textit{should} be used for data fitting. 

We carry out comprehensive data-fitting experiments on GPT-4o-mini (a representative general-purpose close-weight LLM),\footnote{Among many close-weight commercial LLMs, GPT-4o-mini is one of the most cost-effective flagship LLMs that support both completion and fine-tuning.} Llama-3-8B-instruct (a representative general-purpose open-weight LLM), as well as TabPFN \cite[currently best-performing tabular foundation model specifically trained for data-fitting tasks,][]{hollmann2025accurate}. We perform LLM-based data fitting via both in-context learning (ICL) and supervised fine-tuning (SFT). For GPT-4o-mini, we use OpenAI's batch processing API for ICL and fine-tuning API for SFT. For Llama-3-8B-instruct, we download the model from Hugging Face\footnote{Source: \url{https://huggingface.co/meta-llama/Meta-Llama-3-8B-Instruct}.} and carry out ICL using resources provided by Minnesota Supercomputing Institute. For general-purpose LLMs (GPT-4o-mini and Llama-3-8B-instruct), we follow standard practices in LLM prompting to represent data instances in a natural language format (e.g., "A data point has X0 [VAL]..."). We also consider an alternative data format (i.e., JSON) as one task-irrelevant variation. The detailed prompts and procedures for both ICL and SFT are reported in \textit{SI Appendix, ``LLM Prompts and Prediction Procedures" Section}. For the tabular foundation model (TabPFN), we use its official Python implementation to perform data fitting via ICL. Given the objective nature of the task, we use greedy decoding for all LLMs (i.e., selecting the next token to be the one receiving highest predicted probability among the candidate tokens). This is done by setting LLM generation temperature to 0 and disabling random token sampling whenever possible. We also fix the random seed during SFT of LLMs as well as the training of all tabular supervised learning models to ensure reproducibility and fair comparisons across different models. All models are evaluated on the same 1,000 testing instances.

Due to the non-trivial monetary and computational costs of LLM usage, we first perform data fitting and evaluations with GPT-4o-mini and Llama-3-8B-instruct on a single (fixed) set of training and testing data (and report the corresponding results). We then repeat data fitting with these LLMs for a second time, in order to test whether the observed prediction sensitivity is caused by task-irrelevant variations or by some inherent randomness in LLMs' token generation. Furthermore, for analyses involving TabPFN, since the tabular foundation model can be run locally and efficiently, we repeat the data-fitting experiments 100 times to obtain statistical evidence of prediction sensitivity. 

Replication materials can be accessed at \url{https://github.com/heziiiiiiiii/LLMs-for-data-fitting}.

\section{LLMs Can Achieve Competitive Prediction Accuracy}

\subsection{Performance Level} Starting with GPT-4o-mini, we evaluate the LLMs' predictive performance level as compared to six widely-used tabular supervised learning techniques, including linear regression (the oracle model in this case), LASSO regression, $k$-NN, Support Vector Regression (SVR), Random Forest, and Multi-Layer Perceptron (MLP). Each supervised technique is trained on the same training data of 4,000 instances, with hyper-parameters tuned via 5-fold cross-validation. For reference, we also include a naive benchmark, namely using the mean of outcome variable on the training data as (constant) prediction. For GPT-4o-mini under SFT, we use all 4,000 training instances for fine-tuning. Under ICL, following the common practice in few-shot prompting, we include either 10 or 20 instances sampled from training data into the task prompt (denoted as ICL10 and ICL20). We also explore an ICL prompt with 500 instances (denoted as ICL500), representing a large-scale setting that is still within the context window of GPT-4o-mini. We evaluate predictive performance on the testing data using the standard Mean Absolute Error (MAE) metric. The results are shown in Table \ref{tab:performance_base}.

\begin{table}[!ht]
    \centering
    \caption{MAE Comparisons between LLM and Supervised Learning Techniques (GPT-4o-mini)}
    \label{tab:performance_base}
    \begin{tabular}{r|c|c}
    \toprule
         & $X_i \sim N(0,1)$ & $X_i \sim \exp(1)$ \\
    \hline
    Naive Benchmark  &  1.5331  &  1.5869  \\
    Linear Regression (Oracle) &  0.0792  &  0.0791  \\
    LASSO  &  0.0825  &  0.0827  \\
    SVR  &  0.0792  &  0.0791  \\
    Random Forest  &  0.4837  &  0.4849  \\
    $k$-NN  &  0.5414  &  0.5416  \\
    MLP  &  0.0807  &  0.0803  \\
    \midrule
    ICL10  &  0.9246  &  0.8749  \\
    ICL20  &  0.9283  &  0.7717  \\
    ICL500  &  0.9315  &  0.7830  \\
    SFT  &  0.1219  &  0.1313  \\
    \bottomrule
    \end{tabular}    
\end{table}

Not surprisingly, linear regression and other tabular supervised learning techniques that can sufficiently approximate a linear relationship (LASSO, SVR, and MLP) all perform quite well on this task, whereas Random Forest and $k$-NN perform comparatively worse. More importantly, LLM-based data fitting achieves respectable performance. With SFT, LLM outperforms Random Forest and $k$-NN and comes somewhat close to the performance of the other techniques. With ICL, even with only 10 or 20 examples (substantially smaller ``training" size than the other models), it still meaningfully outperforms the naive benchmark, indicating that LLM can conduct few-shot learning to some degree in a prediction task. Further increasing the number of examples in ICL to 500 does not necessarily improve predictive performance.\footnote{We further consider a few variations to the above baseline setting, including different numbers of variables, variable scales, and the presence of outliers in data. Overall, we find that LLMs can achieve reasonable performance levels across these variations, especially with SFT.}

%\subsection*{Performance Level under other Settings} We further consider a few variations to the above baseline setting, including different numbers of variables, variable scales, and the presence of outliers in data. These variations change the inherent ``difficulty" of the prediction task, and can have differential impact on both supervised techniques and LLMs (e.g., linear regression is unaffected by variable scales but sensitive to outliers, but the reverse is true for $k$-NN). Overall, we find that LLMs achieve reasonable performance levels across these variations (especially with SFT). LLMs also behave similarly as random forest and $k$-NN in that their performance are less affected by outliers, but more affected by having more input variables or having variables of different scales. Given our focus on understanding LLMs' prediction robustness, we include the results of these experiments in \textit{SI Appendix, Section 2}.

\section{LLMs Have Poor Prediction Robustness with Respect to Task-Irrelevant Variations}

Achieving advantageous predictive performance is not the only requirement for data-fitting techniques in practice. We now evaluate LLMs' task-irrelevant prediction sensitivity -- whether (and how much) LLMs' predictions for the same data instances would vary in the presence of  \textit{task-irrelevant variations} to input data, i.e., when we apply changes that do not alter the underlying prediction task in any meaningful way. More specifically, we consider five types of task-irrelevant variations, described as follows:

\begin{itemize}
    \item \textbf{Variable Names}: change the names of input variables from ``X0", ``X1", $\ldots$, ``X9" to ``First\_Variable", ``Second\_Variable", $\ldots$, ``Tenth\_Variable";
    \item \textbf{Variable Order}: randomly shuffle the order of variables (maintaining the same variable order across data instances);
    \item \textbf{Number of Digits after decimal point}: round each numerical value from 15 digits (the default precision of a double float number in Python) to 10 digits.\footnote{While doing so technically reduces the precision of input variables, the applied changes are highly negligible as compared to how we measure and report predictive performance (4 digits after decimal point). We therefore still consider it as a task-irrelevant variation.}
    \item \textbf{Format}: change the natural language format of data instance representation (e.g., ``A data point has X1 0.1269...") to a JSON key-value format.
    \item \textbf{Row Order}: for ICL, reverse the order of the few-shot examples in LLM prompts; for SFT, use a different random seed (which would lead to a different ordering of data instances during fine-tuning).
\end{itemize}

Note that these changes do not meaningfully alter the actual data or the data-fitting task, and tabular supervised learning techniques would not be materially affected by them. However, we observe that LLM-generated predictions have poor robustness across all of these changes, and the sensitivity can be quite high under some settings. In particular, let $MAE_{base}$ and $MAE_{change}$ respectively denote the predictive performance before and after applying task-irrelevant changes, we define the prediction sensitivity metric as
\begin{equation}
\label{eq:sensitivity}
\text{\%Sensitivity} = \frac{|MAE_{base} - MAE_{change}|}{MAE_{base}} \times 100\%
\end{equation}
and we report the results in Table \ref{tab:performance_sensitivity}.

\begin{table}[!ht]
    \centering
    \caption{LLM Prediction Sensitivity with Respect to Task-Irrelevant Variations (GPT-4o-mini)}
    \label{tab:performance_sensitivity}
    \begin{tabular}{r|c|c|c|c}
    \toprule
        & \multicolumn{4}{c}{$X_i \sim N(0,1)$} \\
    \hline
      &  ICL10  &  ICL20  &  ICL500  &  SFT  \\
    \hline
    Variable Names  &  26.06\%  &  28.00\%  &  29.47\%  &  2.19\%  \\
    Variable Order  &  3.46\%  &  0.08\%  &  11.68\%  &  0.07\%  \\
    Number of Digits  &  1.64\%  &  5.74\%  &  0.26\%  &  0.07\%  \\
    Format  &  5.70\%  &  7.86\%  &  5.78\%  &  23.78\%  \\
    Row Order  &  34.00\%  &  25.87\%  &  9.77\%  &  5.12\% \\
    \midrule
      &  \multicolumn{4}{c}{$X_i \sim \exp(1)$} \\
    \hline
      &  ICL10  &  ICL20  &  ICL500  &  SFT  \\
    \hline
    Variable Names  &  82.06\%  &  39.68\%  &  5.76\%  &  2.39\%  \\
    Variable Order  &  1.37\%  &  8.64\%  &  10.53\%  &  3.18\%  \\
    Number of Digits  &  0.33\%  &  8.52\%  &  11.95\%  &  5.19\%  \\
    Format  &  3.40\%  &  0.43\%  &  4.27\%  &  3.59\%  \\
    Row Order  &  69.19\%  &  32.05\%  &  1.95\%  &  8.34\%  \\
    \bottomrule
    \end{tabular}    
\end{table}

All task-irrelevant variations lead to nontrivial sensitivity in LLM predictions, which can be as high as 82\% under ICL and 24\% under SFT. Worse yet, we do not observe any clear patterns across different settings -- SFT does not necessarily lead to more robust predictions than ICL; ICL with more in-context examples does not lead to greater robustness; and sensitivity also varies with the statistical distribution of input variables. In other words, LLM-based data fitting and predictions not only suffer from poor robustness with respect to fundamentally task-irrelevant variations, the degree of sensitivity is also hard to anticipate a priori.

In addition to applying each task-irrelevant change separately, we also consider the compounding impact of more than one task-irrelevant change. As an illustrative example, we examine a combination of ``Variable Order" and ``Format", where the training data are represented in a JSON key-value format \textit{and} each instance has randomized variable order. This poses no effect on tabular supervised learning techniques, but we again observe clear prediction sensitivity in LLM. With exponentially distributed covariates, GPT-4o-mini performance varies by 8.44\% under ICL10 setting, 1.48\% under ICL20 setting, 6.51\% under ICL500 setting, and 0.33\% under SFT setting. Furthermore, sensitivity due to the combined change differs from that of the individual changes in an unpredictable manner (bigger in some settings and smaller in others). 

\subsection{Prediction Sensitivity Cannot Be Attributed to Randomness} Data fitting and prediction algorithms often involves some randomness. Among tabular supervised learning techniques, random forest is a representative example. A random forest consists of individual decision trees built on randomly selected training instances and feature sets. As a result, applying task-irrelevant variations to the data (e.g., changing the order of variables or rows) may also alter the trained model and its predictions. This gives rise to a natural question: does a random forest exhibit similar degrees of task-irrelevant prediction sensitivity as LLMs? We evaluate this by applying the ``Variable Order", ``Number of Digits", and ``Row Order" variations\footnote{The ``Variable Names" and ``Format" variations are not applicable in a tabular data fitting context, and therefore would not cause any prediction sensitivity by definition.} and then report the average prediction sensitivity of a random forest model across 100 repetitions (each with different datasets simulated from the same data-generation process). Under exponentially distributed covariates, we find that the prediction sensitivity is 0.01\% under ``Variable Order" variation, 0.01\% under ``Number of Digits" variation, and 0.15\% under ``Row Order" variation, all of which are \textit{one to two orders of magnitude smaller} than GPT-4o-mini's corresponding prediction sensitivity. The results under normally distributed covariates are qualitatively consistent. Furthermore, we re-train a random forest model on the exact same data (without applying any task-irrelevant variation) but set a different random seed. We find that average prediction sensitivity caused by varying random seed is 0.14\%, which is not significantly different from the sensitivity associated with applying ``Row Order" variation ($p = 0.66$) and significantly greater than the sensitivity associated with applying ``Variable Order" or ``Number of Digits" variations ($p < 0.001$). In other words, a random forest's task-irrelevant prediction sensitivity \textit{does not exceed} what is attributable to randomness in its data-fitting process.

Besides random forest, MLP is another technique that can exhibit task-irrelevant prediction sensitivity. We repeat the above analysis with MLP and find the prediction sensitivity to be 0.01\% under ``Variable Order" variation, which is not significantly different from sensitivity due to simply varying random seed (0.01\%, $p = 0.99$). Prediction sensitivity under ``Number of Digits" and ``Row Order" variations are negligible ($1.38 \times 10^{-6}$\% and $1.30 \times 10^{-10}$\% respectively) and significantly lower than the random seed-induced sensitivity ($p < 0.001$).

In stark contrast, we find that LLMs' task-irrelevant prediction sensitivity \textit{exceeds what can be explained by randomness}. LLM-based data fitting also involves randomness, primarily manifested as randomness in token generation. Moreover, such randomness is not as easy to control as in random forests -- even though we set LLM generation temperature to 0 in all our experiments, re-running the exact same prompt may still generate slightly different predictions, due to ties in predicted logit values of candidate tokens and the potential randomness in the LLM's tie-breaking approach.\footnote{For instance, if two candidate tokens (e.g., ``123" and ``124") share the same predicted logit value during one inference step, some LLMs would {\em randomly} choose one of these two candidates as the output token. OpenAI API does not offer precise control over this randomness.} Next, we perform statistical tests to ascertain that the prediction sensitivity we observe cannot be attributed to such randomness in token generation.

Specifically, to assess the ``baseline" prediction sensitivity that is purely caused by token generation randomness, we re-run each setting (both ICL and SFT) for a second time, without applying any task-irrelevant changes. For a given testing instance $\boldsymbol{x}$, we would obtain two predictions from the two identical runs, $LLM(\boldsymbol{x})$ and $LLM'(\boldsymbol{x})$, that could potentially be slightly different. For the same testing instance under a given task-irrelevant change, let $LLM^{change}(\boldsymbol{x})$ denote the LLM's prediction. Next, we compute prediction variation due to randomness $\Delta_{rand} = |LLM(\boldsymbol{x}) - LLM'(\boldsymbol{x})|$ as well as prediction variation due to the task-irrelevant change $\Delta_{change} = |LLM(\boldsymbol{x}) - LLM^{change}(\boldsymbol{x})|$. 

If a task-irrelevant change leads to prediction sensitivity that cannot be simply explained away by token generation randomness, we would expect $\Delta_{change}$ to significantly exceed $\Delta_{rand}$ on average. We check this by performing one-sided paired $t$-tests comparing the two quantities across 1,000 testing instances. We find that the differences are statistically significant at 0.001 level for all but two settings (where the prediction sensitivity is small); the only exceptions are SFT with normally distributed variables under variable name change and number of digits change ($p = 0.059$ and $p = 0.091$, both significant at 0.1 level).

\subsection{Prediction Sensitivity Cannot Be Avoided by Prompting} We further demonstrate that prediction sensitivity cannot be prevented by simply telling the LLM not to be sensitive with respect to task-irrelevant variations. Let's consider the Row Order variation as an illustrative example. Under the ICL setting (where training examples are part of the prompt), we insert an extra instruction after the ICL examples that states ``\textit{The $K$ examples presented above are not in any particular order.}" This is intended to signal to the LLM that the specific ordering of training examples has no bearing on the data-fitting task. 

In Table \ref{tab:performance_prompt}, we present the predictive performance and sensitivity associated with three configurations: (i) Baseline (i.e., the original ICL prompting); (ii) Row Order (i.e., applying the row order change); (iii) Row Order $+$ Extra Instruction. We report results based on exponentially distributed variables (the results based on normally distributed variables are qualitatively consistent).

\begin{table}[!ht]
    \centering
    \caption{Predictive Performance and Sensitivity with Extra Prompt Instruction ($X_i \sim \exp(1)$, prediction sensitivity compared to Baseline reported in parentheses)}
    \label{tab:performance_prompt}
    \begin{tabular}{r|c|c|c}
    \toprule
      &  ICL10  &  ICL20  & ICL500   \\
    \hline
    (i) Baseline MAE &  0.8749  &  0.7717 &  0.7830  \\
    \midrule
    %(ii) Baseline $+$ Extra Inst.  &  0.8507  &  0.7530  &  0.8113 \\
    %  &   (2.77\%)  & (2.42\%)  &  (3.61\%) \\
    (ii) Row Order  &  1.4803  &  1.0190 &  0.7983 \\
      &  (69.19\%)  &  (32.05\%) &  (1.95\%) \\
    (iii) Row Order $+$ Extra Inst.  &  1.2605 & 0.8900 &  0.8278 \\
      &  (44.06\%) & (15.33\%) &  (5.71\%) \\
    \bottomrule
    \end{tabular}    
\end{table}

The extra prompt instruction is clearly ineffective in countering prediction sensitivity induced by row order change. In configuration (iii), we still observe nontrivial sensitivity (up to 44.06\%). Moreover, comparing configurations (ii) and (iii), we can see that inserting the extra instruction, which is intended to improve prediction robustness, can actually lead to \textit{higher} prediction sensitivity in the ICL500 setting. 

In general, prompt engineering alone cannot resolve LLMs' prediction sensitivity with respect to task-irrelevant variations. Even if one can somehow anticipate the variations to guard against ahead of time and insert corresponding prompt instructions, doing so can end up increasing (rather than decreasing) task-irrelevant prediction sensitivity.

\section{Replication with an Open-Weight LLM and Mechanism Exploration}

We repeat the performance level and sensitivity analyses with a flagship open-weight model, Llama-3-8B-instruct. The open-weight nature of this LLM also provides a unique opportunity to explore the underlying reason for LLMs' lack of prediction robustness, by examining the internal attention scores. We focus on the ICL10 and ICL20 settings here, because ICL500 or SFT with this model require extensive computing resources that are beyond what is available to us. We report results under exponentially distributed variables (the results under normally distributed variables are qualitatively similar).

\subsection{\textbf{Results}} Table \ref{tab:performance_llama} summarizes both the predictive performance and the prediction sensitivity of Llama-3-8B-instruct. Overall, it achieves even better predictive performance than GPT-4o-mini under the ICL settings. However, like GPT-4o-mini, its predictions are sensitive with respect to task-irrelevant variations. Note that we set the Llama model to use greedy decoding without random token sampling, so re-running the same setting yields exactly the same set of predictions; thus, the observed prediction sensitivity still cannot be attributed to token generation randomness. Also, the degrees of prediction sensitivity are much higher than that of random forest or MLP, further indicating that Llama's task-irrelevant prediction sensitivity is not just driven by randomness.

\begin{table}[!ht]
    \centering
    \caption{Performance Level and Sensitivity of Llama-3-8B-instruct ($X_i \sim \exp(1)$)}
    \label{tab:performance_llama}
    \begin{tabular}{r|c|c}
    \toprule
      &  ICL10  &  ICL20    \\
    \hline
    Baseline MAE  &  0.7911  &  0.6757  \\
    \hline
    Variable Names  &  2.82\%  & 0.80\%  \\
    Variable Order  &  5.28\%  & 9.75\%  \\
    Number of Digits  &  6.38\%  &  5.49\%   \\
    Format  &  0.64\%  &  3.32\%   \\
    Row Order  &  9.68\%  &  3.67\%  \\
    \bottomrule
    \end{tabular}    
\end{table}

\subsection{Attention-Based Mechanism Exploration} The black-box nature of LLMs makes it challenging to pin down exactly why they generate some specific tokens over others. Several research studies have explored the use of attention scores to derive possible explanations for certain LLM behaviors \citep[e.g.,][]{elhage2021mathematical,zhou2024unibias,kamath2025tracing}. For auto-regressive LLMs, attention scores are crucial in determining which token to generate next \cite{vaswani2017attention}. We investigate the {\em cross-attention} from the tokens that make up the prediction value to the tokens that represent the ICL examples, in order to probe the potential mechanism behind LLMs' prediction sensitivity.

More concretely, consider an ICL prompt with $K$ examples included in it. Let $P_k$ denote the set of tokens that make up the $k$-th example, let $O$ denote the output tokens that make up the LLM's prediction value. For each pair of tokens $o \in O$ and $p \in P_k$, let $Attn(o,p)$ denote the cross-attention score that token $o$ pays to token $p$.\footnote{Llama-3-8B-instruct has 32 attention heads. $Attn(o,p)$ is the average attention score across all attention heads.} We can define a ``row attention" that quantifies the amount of attention from output tokens to one ICL example:
\begin{equation}
\label{eq:row_attn}
RowAttn(k) = \frac{1}{|O|} \sum_{o \in O} \sum_{p \in P_k} Attn(o,p)
\end{equation}
We normalize it by token count in the output so that row attention is not skewed by output length. Moreover, $P_k$ can be broken down to three subsets of tokens -- those representing variable names, those representing variable values, and other miscellaneous tokens (blank space, comma, etc.). Let $PName_{ik}$ and $PVal_{ik}$ respectively denote the variable name tokens and variable value tokens corresponding to the $i$-th variable ($i \in \{1, \ldots, 10\}$) in the $k$-th example. We further define  ``variable name attention" and ``variable value attention" as follows:
\begin{equation}
\label{eq:var_attn}
\begin{split}
    NameAttn(i) &= \frac{1}{|O|} \sum^K_{k=1} \sum_{o \in O} \sum_{p \in PName_{ik}} Attn(o,p) \\
    ValAttn(i) &= \frac{1}{|O|} \sum^K_{k=1} \sum_{o \in O} \sum_{p \in PVal_{ik}} Attn(o,p) \\
\end{split}
\end{equation}

\begin{figure}[!ht]
    \centering
    \includegraphics[width=0.6\linewidth]{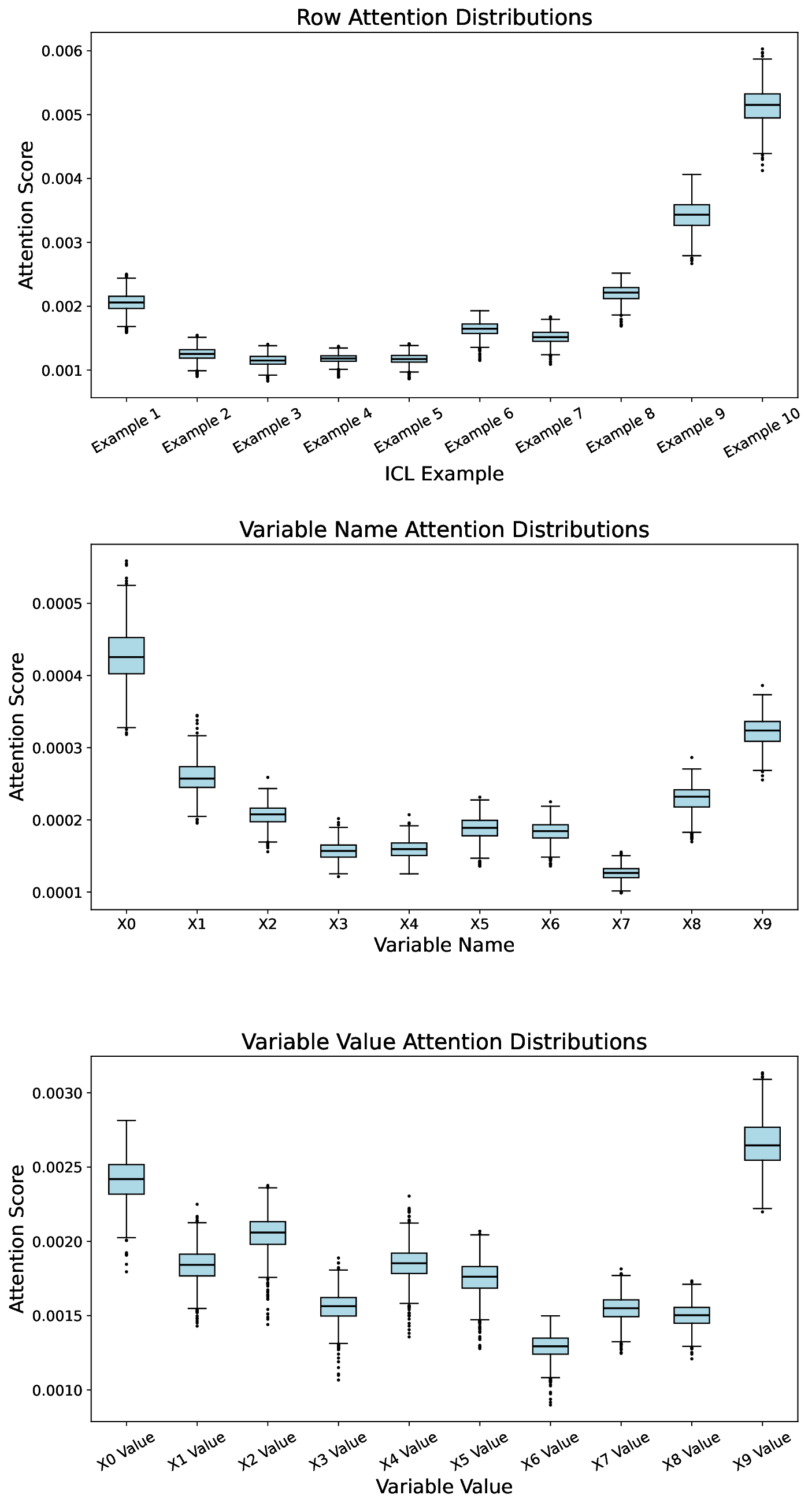}
    \caption{Attention Distributions by Rows, Variable Names, and Variable Values (Llama-3-8B-instruct under ICL10 setting)}
    \label{fig:attention_icl10}
\end{figure}

In a canonical data-fitting task -- learning a model that maps input values to target predictions -- there should not \textit{a priori} be any differential focus on specific rows / variable names / variable values (e.g., variable names are not even part of the canonical definition of a data-fitting task). In other words, the expectation would be that $RowAttn(k)$ is roughly the same across $k$; $NameAttn(i)$ and $ValAttn(i)$ should be roughly the same across $i$. In fact, these types of expected attention ``regularities" are explicitly engineered into special-purpose tabular foundation models \citep[e.g., TabPFN,][as will be discussed later]{hollmann2025accurate}. However, as can be seen in the results in Figure \ref{fig:attention_icl10}, we observe a ``U-shaped" attention pattern: rows at the beginning or end of the ICL examples receive more attention than other rows, and variable names / values at the beginning or end of a given example receive more attention than those at other positions. Such attention distributions are statistically significant -- a Kruskal-Wallis test shows that all three attention distributions are different from being uniform ($p < 0.001$ in all three cases).

We observe similar ``U-shaped" attention distributions under the ICL20 setting (with 20 ICL examples that are different from those in ICL10). This further indicates that the ``U-shaped" pattern is associated with specific positions in the ICL prompt rather than the training examples we use. See \textit{SI Appendix, ``Additional Analyses of Attention Distributions" Section} for the detailed results.

The uneven attention distributions represent a plausible explanation for LLMs' prediction sensitivity, because training examples and variables that happen to occupy the beginning / end positions in a prompt end up having larger influence in determining the output tokens. To illustrate this, we repeat the ICL10 experiment but shuffle the variable order. The ``U-shaped" attention distributions persist, and variables that show up at beginning / end of each example by random chance now receive greater attentions from the output tokens. We include the results in \textit{SI Appendix, ``Additional Analyses of Attention Distributions" Section}.

Our findings are reminiscent of the phenomena of ``position bias" \citep[][where positions of in-context examples in user prompts can influence few-shot learning performance]{zhao2021calibrate,cobbina2025show}, ``attention sink" \citep[][where attention scores toward initial tokens tend to be stronger even if those tokens are not semantically meaningful]{xiao2023efficient}, and ``lost in the middle" \citep[][where LLMs' information retrieval performance tends to be worse when relevant information is in the middle part of a long context]{liu2023lost}. Meanwhile, the attention pattern has a richer, more nuanced structure in our context, as the ``U-shaped" distributions manifest both across few-shot examples and within each example (across variable names / values).

\section{Specialized Tabular Foundation Models Are Not Immune to Prediction Sensitivity}

Next, we extend our analyses to tabular foundation models specifically designed for data-fitting and prediction tasks. TabPFN \cite{hollmann2025accurate} represents a state-of-the-art tabular foundation model that has outperformed supervised learning techniques on a wide variety of benchmarking datasets. Importantly, the design of TabPFN recognizes that data fitting should be invariant with respect to variable order and row order, and explicitly tries to facilitate this by implementing a ``two-way attention" mechanism where each tabular cell attends to other features in the same row as well as to other values under the same column. 

TabPFN can carry out ICL with up to 500 input variables and 10,000 data instances. Therefore, in addition to the ICL10 / ICL20 / ICL500 settings, we also put the entire training dataset (4,000 training instances, denoted as ICL4000) to TabPFN and obtain predictions on the same 1,000 testing instances. Note that TabPFN does not take variable names and requires a standard dataframe format, so we focus on the other three task-irrelevant variations, namely ``Variable Order", ``Number of Digits", and ``Row Order". We use the official \texttt{TabPFNRegressor} implementation with default parameters. The computational efficiency of TabPFN allows us to repeat the evaluations 100 times, each with different training / testing datasets simulated from the same data-generation process. We report the mean and standard deviation of performance level and prediction sensitivity in Table \ref{tab:performance_tabpfn} for exponentially distributed variables (normally distributed variables yield similar results).

\begin{table}[!ht]
    \centering
    \caption{Performance Level and Sensitivity of TabPFN ($X_i \sim \exp(1)$, standard errors of MAE and prediction sensitivity are reported in parentheses)}
    \label{tab:performance_tabpfn}
    \begin{tabular}{r|c|c|c|c}
    \toprule
       & ICL10 & ICL20 & ICL500 &  ICL4000 \\
    \hline
    Baseline MAE  & 1.0499 & 0.7036 & 0.1038  &  0.0879  \\
      &  (0.2773)  &  (0.1920)  &  (0.0059)  &  (0.0034) \\
    \hline
    Variable Order & 2.83\% & 3.04\% & 1.34\%  &  0.75\% \\
      &  (0.0295)  &  (0.0280)  &  (0.0109)  &  (0.0055) \\
    Number of Digits  & 1.16\% & 1.64\% & 0.58\% &  0.22\% \\
      &  (0.0123)  &  (0.0140)  &  (0.0058)  &  (0.0019) \\
    Row Order  & 0.76\% & 1.36\% & 0.46\%  & 0.20\% \\
      &  (0.0083)  &  (0.0108)  &  (0.0043)  &  (0.0014) \\
    \bottomrule
    \end{tabular}
\end{table}

TabPNF achieves impressive predictive performance -- with all 4,000 training examples, its average MAE clearly outperforms general-purpose LLMs (GPT-4o-mini and Llama-3-8B-instruct) under both ICL and SFT. Nevertheless, it is not immune to task-irrelevant variations, with average prediction sensitivity ranging from 0.20\% to 3.04\%, which is still higher than that of canonical tabular machine learning techniques, including random forest or MLP. Moreover, like the Llama model, re-running TabPFN with the same prompt yields exactly the same set of predictions, so the observed prediction sensitivity again cannot be attributed to token generation randomness.

Based on the above results, one might be tempted to conclude that the degree of prediction sensitivity under TabPFN is relatively small, i.e., perhaps not substantial enough to warrant further concern. We argue, however, that this interpretation misses the central issue -- that task-irrelevant variations should not cause \textit{any} prediction changes (beyond what can be explained by randomness in the data-fitting process) for a tool that possesses basic competence in data-fitting tasks. As mentioned earlier, this could be viewed as essentially analogous to having a calculator that produces different results when adding up the same set of numbers in a different order; even if the differences appear ``small", hardly anyone would be comfortable using it as a competent calculation tool. 

The fact that TabPFN is not immune to task-irrelevant variations points to a deeper issue. Despite TabPFN's explicit architectural choices to facilitate the invariance with respect to variable order and row order, it cannot fully achieve the said invariance. In contrast, tabular supervised learning techniques are, \textit{by design}, either completely invariant to these changes or exhibit minor sensitivity due to randomness (e.g., in random forest or MLP). We argue that this highlights a fundamental challenge with using current LLMs for data fitting: although robustness with respect to some task-irrelevant variations could be facilitated to a limited extent with special-purpose tabular foundation models, it generally cannot be guaranteed.

\section{Conclusions}
Modern LLMs have demonstrated impressive performance across a wide array of settings. As a result, they are increasingly being used as ``general-purpose" AI tools, beyond the typical language-oriented tasks. One such task is data fitting and prediction, where LLMs have been shown to perform competitively as or even outperform common tabular supervised learning techniques via in-context learning or supervised fine-tuning \citep{gruver2023large,vacareanu2024words,alnegheimish2024large,dong2024can}. 

We argue that just because LLMs \textit{can} be used for data fitting does not mean they \textit{should} be. Through extensive numerical experiments, we demonstrate that current LLMs have poor prediction robustness with respect to task-irrelevant variations. Altering variable names, order of variables, order of training examples, or inconsequential details of data representation -- all of which are completely irrelevant to the underlying data-fitting task -- can significantly change the LLMs' predictions on the same data. 

Such prediction sensitivity persists across different learning methods (in-context learning or supervised fine-tuning) as well as different LLMs (general-purpose close- or open-weight LLMs and special-purpose tabular foundation models), and cannot simply be attributed to randomness in token generation. Instead, we provide descriptive evidence that the output tokens are paying more attention (a) to training examples that happen to occupy the beginning or end positions of a prompt and (b) to variable names / values that happen to occupy the beginning or end positions within each training example, even if those positions are not more important a priori. The non-uniform attention distributions provide a plausible explanation behind spurious sensitivity in LLMs' predictions.

We believe the absence of even a basic level of prediction robustness brings several challenges. First, it introduces additional complications to the (already difficult) challenge of LLM interpretability \citep{nanda2023progress,conmy2023towards}. Why an LLM produces a certain prediction depends not only on the underlying data but also on specific, task-irrelevant variations of the prompt. Second, it raises safety, reliability, and trust concerns when LLM-based data fitting and prediction systems are deployed in practice. For instance, if predictions on the same data can be easily altered by changing some variable names, how much would we want to trust those predictions, let alone to make (potentially high-stakes) decisions based on them? Last but not least, people familiar with tabular supervised learning techniques typically take for granted that task-irrelevant variations would not affect their models or predictions beyond what can be explained by randomness (e.g., due to having different random seeds); in contrast, the idiosyncratic choices of LLM users related to data representation can greatly change the results of the same analyses. As one of the consequences, it can further worsen the reproducibility crisis in science \citep{baker20161,youyou2023discipline} due to the ever increasing demand for data analysis, data fitting, and predictive modeling in scientific inquiry and the increasing use of LLMs for this purpose. 

Our study resonates with recent work that reveals LLM limitations in imitating human behaviors \citep[][where the language used in prompting can significantly alter LLM behavior patterns]{gao2025take} and mathematical reasoning \citep[][where changes to person names appearing in a math question clearly affects LLM performance]{mirzadeh2024gsm}. In all of these studies, the tasks of interest are commonly understood to involve learning proper abstractions and applying principled procedures, which is incompatible with observations that innocuous, task-irrelevant modifications end up causing drastic changes in LLM outputs. Taken together, this line of inquiry highlights an important challenge of current LLMs, namely their inability to distinguish what is completely irrelevant in a given task. As discussed earlier, we believe such limitations cast serious doubt on whether current LLMs, as a learning paradigm, exhibit ``competence" as a problem-solving tool. Addressing such limitations represents an important direction for future work.

%%%%%%%%%Bibliography%%%%%%%%%%
\bibliographystyle{apalike}%%%%
\bibliography{arxiv_ref}%%%

\begin{thebibliography}{}

\bibitem[Akhauri et~al., 2025]{akhauri2025performance}
Akhauri, Y., Lewandowski, B., Lin, C.-H., Reyes, A.~N., Forbes, G.~C., Wongpanich, A., Yang, B., Abdelfattah, M.~S., Perel, S., and Song, X. (2025).
\newblock Performance prediction for large systems via text-to-text regression.
\newblock {\em arXiv preprint arXiv:2506.21718}.

\bibitem[Alnegheimish et~al., 2024]{alnegheimish2024large}
Alnegheimish, S., Nguyen, L., Berti-Equille, L., and Veeramachaneni, K. (2024).
\newblock Large language models can be zero-shot anomaly detectors for time series?
\newblock {\em arXiv preprint arXiv:2405.14755}.

\bibitem[Baker, 2016]{baker20161}
Baker, M. (2016).
\newblock 1,500 scientists lift the lid on reproducibility.

\bibitem[Brand et~al., 2023]{brand2023using}
Brand, J., Israeli, A., and Ngwe, D. (2023).
\newblock Using llms for market research.
\newblock {\em Harvard business school marketing unit working paper}, (23-062).

\bibitem[Broadbent, 1958]{broadbent1958perception}
Broadbent, D.~E. (1958).
\newblock {\em Perception and communication}.
\newblock Elsevier.

\bibitem[Brown et~al., 2020]{brown2020language}
Brown, T., Mann, B., Ryder, N., Subbiah, M., Kaplan, J.~D., Dhariwal, P., Neelakantan, A., Shyam, P., Sastry, G., Askell, A., et~al. (2020).
\newblock Language models are few-shot learners.
\newblock {\em Advances in neural information processing systems}, 33:1877--1901.

\bibitem[Cobbina and Zhou, 2025]{cobbina2025show}
Cobbina, K. and Zhou, T. (2025).
\newblock Where to show demos in your prompt: A positional bias of in-context learning.
\newblock {\em arXiv preprint arXiv:2507.22887}.

\bibitem[Conmy et~al., 2023]{conmy2023towards}
Conmy, A., Mavor-Parker, A., Lynch, A., Heimersheim, S., and Garriga-Alonso, A. (2023).
\newblock Towards automated circuit discovery for mechanistic interpretability.
\newblock {\em Advances in Neural Information Processing Systems}, 36:16318--16352.

\bibitem[Das et~al., 2024]{das2024decoder}
Das, A., Kong, W., Sen, R., and Zhou, Y. (2024).
\newblock A decoder-only foundation model for time-series forecasting.
\newblock In {\em Forty-first International Conference on Machine Learning}.

\bibitem[Dong et~al., 2024]{dong2024can}
Dong, M., Huang, H., and Cao, L. (2024).
\newblock Can llms serve as time series anomaly detectors?
\newblock {\em arXiv preprint arXiv:2408.03475}.

\bibitem[Elhage et~al., 2021]{elhage2021mathematical}
Elhage, N., Nanda, N., Olsson, C., Henighan, T., Joseph, N., Mann, B., Askell, A., Bai, Y., Chen, A., Conerly, T., et~al. (2021).
\newblock A mathematical framework for transformer circuits.
\newblock {\em Transformer Circuits Thread}, 1(1):12.

\bibitem[Eloundou et~al., 2024]{eloundou2024gpt}
Eloundou, T., Manning, S., Mishkin, P., and Rock, D. (2024).
\newblock Gpts are gpts: Labor market impact potential of llms.
\newblock {\em Science}, 384(6702):1306--1308.

\bibitem[Gao et~al., 2025]{gao2025take}
Gao, Y., Lee, D., Burtch, G., and Fazelpour, S. (2025).
\newblock Take caution in using llms as human surrogates.
\newblock {\em Proceedings of the National Academy of Sciences}, 122(24):e2501660122.

\bibitem[Giere, 2010]{giere2010explaining}
Giere, R.~N. (2010).
\newblock {\em Explaining science: A cognitive approach}.
\newblock University of Chicago Press.

\bibitem[Goswami et~al., 2024]{goswami2024moment}
Goswami, M., Szafer, K., Choudhry, A., Cai, Y., Li, S., and Dubrawski, A. (2024).
\newblock Moment: A family of open time-series foundation models.
\newblock {\em arXiv preprint arXiv:2402.03885}.

\bibitem[Gruver et~al., 2023]{gruver2023large}
Gruver, N., Finzi, M., Qiu, S., and Wilson, A.~G. (2023).
\newblock Large language models are zero-shot time series forecasters.
\newblock {\em Advances in Neural Information Processing Systems}, 36:19622--19635.

\bibitem[Hollmann et~al., 2025]{hollmann2025accurate}
Hollmann, N., M{\"u}ller, S., Purucker, L., Krishnakumar, A., K{\"o}rfer, M., Hoo, S.~B., Schirrmeister, R.~T., and Hutter, F. (2025).
\newblock Accurate predictions on small data with a tabular foundation model.
\newblock {\em Nature}, 637(8045):319--326.

\bibitem[Hu et~al., 2021]{hu2021lora}
Hu, E.~J., Shen, Y., Wallis, P., Allen-Zhu, Z., Li, Y., Wang, S., Wang, L., and Chen, W. (2021).
\newblock Lora: Low-rank adaptation of large language models.
\newblock {\em arXiv preprint arXiv:2106.09685}.

\bibitem[Huynh et~al., 2025]{huynhimproving}
Huynh, N., Piskorz, J., Berrevoets, J., Luyten, M.~R., and van~der Schaar, M. (2025).
\newblock Improving treatment effect estimation with llm-based data augmentation.
\newblock In {\em 1st ICML Workshop on Foundation Models for Structured Data}.

\bibitem[Isomura et~al., 2024]{isomura2024llmovertab}
Isomura, T., Shimizu, R., and Goto, M. (2024).
\newblock Llmovertab: Tabular data augmentation with language model-driven oversampling.
\newblock {\em Available at SSRN 4821750}.

\bibitem[Jeong et~al., 2024]{jeong2024llm}
Jeong, D.~P., Lipton, Z.~C., and Ravikumar, P. (2024).
\newblock Llm-select: Feature selection with large language models.
\newblock {\em arXiv preprint arXiv:2407.02694}.

\bibitem[Jin et~al., 2023]{jin2023time}
Jin, M., Wang, S., Ma, L., Chu, Z., Zhang, J.~Y., Shi, X., Chen, P.-Y., Liang, Y., Li, Y.-F., Pan, S., et~al. (2023).
\newblock Time-llm: Time series forecasting by reprogramming large language models.
\newblock {\em arXiv preprint arXiv:2310.01728}.

\bibitem[Johnston and Dark, 1986]{johnston1986selective}
Johnston, W.~A. and Dark, V.~J. (1986).
\newblock Selective attention.
\newblock {\em Annual review of psychology}.

\bibitem[Kamath et~al., 2025]{kamath2025tracing}
Kamath, H., Ameisen, E., Kauvar, I., Luger, R., Gurnee, W., Pearce, A., Zimmerman, S., Batson, J., Conerly, T., Olah, C., and Lindsey, J. (2025).
\newblock Tracing attention computation: Attention connects features, and features direct attention.
\newblock {\em Transformer Circuits Thread}.

\bibitem[Liu et~al., 2025]{liu2025large}
Liu, J., Zhang, C., Qian, J., Ma, M., Qin, S., Bansal, C., Lin, Q., Rajmohan, S., and Zhang, D. (2025).
\newblock Large language models can deliver accurate and interpretable time series anomaly detection.
\newblock In {\em Proceedings of the 31st ACM SIGKDD Conference on Knowledge Discovery and Data Mining V. 2}, pages 4623--4634.

\bibitem[Liu et~al., 2023]{liu2023lost}
Liu, N.~F., Lin, K., Hewitt, J., Paranjape, A., Bevilacqua, M., Petroni, F., and Liang, P. (2023).
\newblock Lost in the middle: How language models use long contexts.
\newblock {\em arXiv preprint arXiv:2307.03172}.

\bibitem[Mirzadeh et~al., 2024]{mirzadeh2024gsm}
Mirzadeh, I., Alizadeh, K., Shahrokhi, H., Tuzel, O., Bengio, S., and Farajtabar, M. (2024).
\newblock Gsm-symbolic: Understanding the limitations of mathematical reasoning in large language models.
\newblock {\em arXiv preprint arXiv:2410.05229}.

\bibitem[Nanda et~al., 2023]{nanda2023progress}
Nanda, N., Chan, L., Lieberum, T., Smith, J., and Steinhardt, J. (2023).
\newblock Progress measures for grokking via mechanistic interpretability.
\newblock {\em arXiv preprint arXiv:2301.05217}.

\bibitem[Rosch, 2024]{rosch2024principles}
Rosch, E. (2024).
\newblock Principles of categorization.
\newblock In {\em Cognition and categorization}, pages 27--48. Routledge.

\bibitem[Seedat et~al., 2023]{seedat2023curated}
Seedat, N., Huynh, N., Van~Breugel, B., and Van Der~Schaar, M. (2023).
\newblock Curated llm: Synergy of llms and data curation for tabular augmentation in low-data regimes.
\newblock {\em arXiv preprint arXiv:2312.12112}.

\bibitem[Sui et~al., 2024]{sui2024table}
Sui, Y., Zhou, M., Zhou, M., Han, S., and Zhang, D. (2024).
\newblock Table meets llm: Can large language models understand structured table data? a benchmark and empirical study.
\newblock In {\em Proceedings of the 17th ACM International Conference on Web Search and Data Mining}, pages 645--654.

\bibitem[Tanlamai et~al., 2024]{tanlamai2024generative}
Tanlamai, J., Khern-am nuai, W., and Cohen, M.~C. (2024).
\newblock Generative ai and price discrimination in the housing market.
\newblock {\em Available at SSRN}.

\bibitem[Vacareanu et~al., 2024]{vacareanu2024words}
Vacareanu, R., Negru, V.-A., Suciu, V., and Surdeanu, M. (2024).
\newblock From words to numbers: Your large language model is secretly a capable regressor when given in-context examples.
\newblock {\em arXiv preprint arXiv:2404.07544}.

\bibitem[Vapnik, 2013]{vapnik2013nature}
Vapnik, V. (2013).
\newblock {\em The nature of statistical learning theory}.
\newblock Springer science \& business media.

\bibitem[Vaswani et~al., 2017]{vaswani2017attention}
Vaswani, A., Shazeer, N., Parmar, N., Uszkoreit, J., Jones, L., Gomez, A.~N., Kaiser, {\L}., and Polosukhin, I. (2017).
\newblock Attention is all you need.
\newblock {\em Advances in neural information processing systems}, 30.

\bibitem[Wang et~al., 2024]{wang2024large}
Wang, M., Zhang, D.~J., and Zhang, H. (2024).
\newblock Large language models for market research: A data-augmentation approach.
\newblock {\em arXiv preprint arXiv:2412.19363}.

\bibitem[Weisberg, 2012]{weisberg2012simulation}
Weisberg, M. (2012).
\newblock {\em Simulation and similarity: Using models to understand the world}.
\newblock Oxford University Press.

\bibitem[Xiao et~al., 2023]{xiao2023efficient}
Xiao, G., Tian, Y., Chen, B., Han, S., and Lewis, M. (2023).
\newblock Efficient streaming language models with attention sinks.
\newblock {\em arXiv preprint arXiv:2309.17453}.

\bibitem[Youyou et~al., 2023]{youyou2023discipline}
Youyou, W., Yang, Y., and Uzzi, B. (2023).
\newblock A discipline-wide investigation of the replicability of psychology papers over the past two decades.
\newblock {\em Proceedings of the National Academy of Sciences}, 120(6):e2208863120.

\bibitem[Zhao et~al., 2021]{zhao2021calibrate}
Zhao, Z., Wallace, E., Feng, S., Klein, D., and Singh, S. (2021).
\newblock Calibrate before use: Improving few-shot performance of language models.
\newblock In {\em International conference on machine learning}, pages 12697--12706. PMLR.

\bibitem[Zhou et~al., 2024]{zhou2024unibias}
Zhou, H., Feng, Z., Zhu, Z., Qian, J., and Mao, K. (2024).
\newblock Unibias: Unveiling and mitigating llm bias through internal attention and ffn manipulation.
\newblock {\em Advances in Neural Information Processing Systems}, 37:102173--102196.

\end{thebibliography}
%%%%%%%%%%%%%%%%%%%%%%%%%%%%%%%

\clearpage
\appendix

\section{Details about Recent Work that Uses LLMs for Data-Fitting}

Table \ref{tab:literature} summarizes recent work that uses LLMs for data fitting. 

\begin{table}[!ht]
    \centering
    \begin{tabular}{p{1in}|p{2.5in}|p{2.5in}}
    \toprule
    Citation  &  Data-Fitting Task  &  How LLMs Perform Data-Fitting and Prediction  \\
    \hline
    \cite{tanlamai2024generative}  & Predict property prices based on property features (e.g., number of bedrooms / bathrooms) and location information (e.g., zip code).  &  Zero-shot prediction. The LLM is given the persona of a realtor, and property features are given in a JSON key-value format in user prompt. The LLM is then asked to estimate the property price based on its ``knowledge of similar markets and home values in the area". \\
    \hline
    \cite{gruver2023large,jin2023time,das2024decoder,goswami2024moment}   &  Time series forecasting. Time series data are treated as strings of digits. &  Zero-shot prediction \citep{gruver2023large}, few-shot in-context learning \citep{jin2023time}, or training special-purpose time series foundation models from scratch \citep{das2024decoder,goswami2024moment}.  \\
    \hline
    \cite{vacareanu2024words}   &  Linear and non-linear regressions.  &  In-context learning with 500 examples. \\
    \hline
    \cite{alnegheimish2024large,dong2024can,liu2025large}  & Anomaly detection from time series data, either by predicting the indices of anomalous observations or by conducting time series forecasting to identify deviations from observed values as anomalies.  &  Zero-shot prediction \citep{alnegheimish2024large} and in-context learning \citep{dong2024can,liu2025large}. \\
    \hline
    \cite{brand2023using, seedat2023curated,wang2024large,isomura2024llmovertab}   &  Data augmentation, e.g., imputing missing values based on complete data records or generating new (synthetic) data based on observed data.  &  Zero-shot prediction \citep{brand2023using}, In-context learning \citep{seedat2023curated,wang2024large} and supervised fine-tuning \citep{isomura2024llmovertab}. \\
    \hline
    \cite{huynhimproving}   &  Causal inference, specifically by predicting counterfactual outcomes to support causal effect estimation.  &  In-context learning.  \\
    \hline
    \cite{akhauri2025performance}   &  Regressions, specifically for predicting metric outcomes in large complex systems.  &  In-context learning.  \\
    \bottomrule
    \end{tabular}
    \caption{Details about Recent Work that Uses LLMs for Data-Fitting}
    \label{tab:literature}
\end{table}

\section{LLM Prompts and Prediction Procedures}

We provide details about LLM prompts and prediction procedures used in our data-fitting experiments. For both in-context learning (ICL) and supervised fine-tuning (SFT) approaches, we use 10 variables ($p=10$) as the base configuration.

For ICL with GPT-4o-mini, we employ a structured prompt consisting of two main components: a task description and few-shot examples. The task description instructs the model: \textit{``Your job is to predict the target value based on some features. You will be given 10 features in total, including X0, X1, X2, X3, X4, X5, X6, X7, X8, X9. Please output the target value as a number. It is very important to only output the target number and nothing else."} This is followed by few-shot examples presented in the format: \textit{``You will be given a total of $K$ examples. Here is example $i$: - A data point has X0 [VAL], X1 [VAL], X2 [VAL], X3 [VAL], X4 [VAL], X5 [VAL], X6 [VAL], X7 [VAL], X8 [VAL], X9 [VAL]. The correct target value of this data point is [VAL]."} This pattern is repeated for $K-1$ additional examples, with [VAL] placeholders filled with actual numerical values from the training data. During ICL prediction, each of the 1,000 test instances is presented without their ground-truth target values as user queries to obtain model completions.

For ICL with Llama-3-8B-instruct, we utilize the model's native chat template format with specific system and user message delimiters. The prompt follows the same structure of task description and few-shot examples, supplemented with Llama-3's special tokens to properly delineate the user-assistant conversation flow. The complete prompt is as follows:

~

\noindent \fbox{%
  \parbox{\textwidth}{%
    \texttt{<|start\_header\_id|>system<|end\_header\_id|>} Your job is to predict the target value based on some features. You will be given 10 features in total, including X0, X1, X2, X3, X4, X5, X6, X7, X8, X9. Please output the target value as a number. It is very important to only output the target number and nothing else. \texttt{<|eot\_id|>}

    ~

    \texttt{<|start\_header\_id|>user<|end\_header\_id|>}Predict the target for: X0 [VAL], X1 [VAL], X2 [VAL], X3 [VAL], X4 [VAL], X5 [VAL], X6 [VAL], X7 [VAL], X8 [VAL], X9 [VAL], \texttt{<|eot\_id|>} \texttt{<|start\_header\_id|>assistant<|end\_header\_id|>} [VAL] \texttt{<|eot\_id|>}
    
    (\textit{repeat for the other few-shot examples})

    ~

    \texttt{<|start\_header\_id|>user<|end\_header\_id|>}Predict the target for: X0 [VAL], X1 [VAL], X2 [VAL], X3 [VAL], X4 [VAL], X5 [VAL], X6 [VAL], X7 [VAL], X8 [VAL], X9 [VAL], \texttt{<|eot\_id|>} \texttt{<|start\_header\_id|>assistant<|end\_header\_id|>}

  }%
}

~

The inference is conducted using computational resources from the Minnesota Supercomputing Institute, with each of the 1,000 test instances presented without their ground-truth target values as user queries to obtain model completions.

For SFT with GPT-4o-mini, we construct fine-tuning data points by combing the task description with individual training instances. A total of 4,000 such fine-tuning data points are provided to OpenAI's SFT service to create a specialized fine-tuned model. The resulting fine-tuned model employs the same task description as its system instruction. During prediction with the SFT approach, each of the 1,000 test instances is presented to the fine-tuned model as a user query to generate predictions.

\section{Additional Analyses of Attention Distributions}

\begin{figure}[!ht]
    \centering
    \begin{minipage}{0.49\textwidth}
        \centering
        \includegraphics[width=\linewidth]{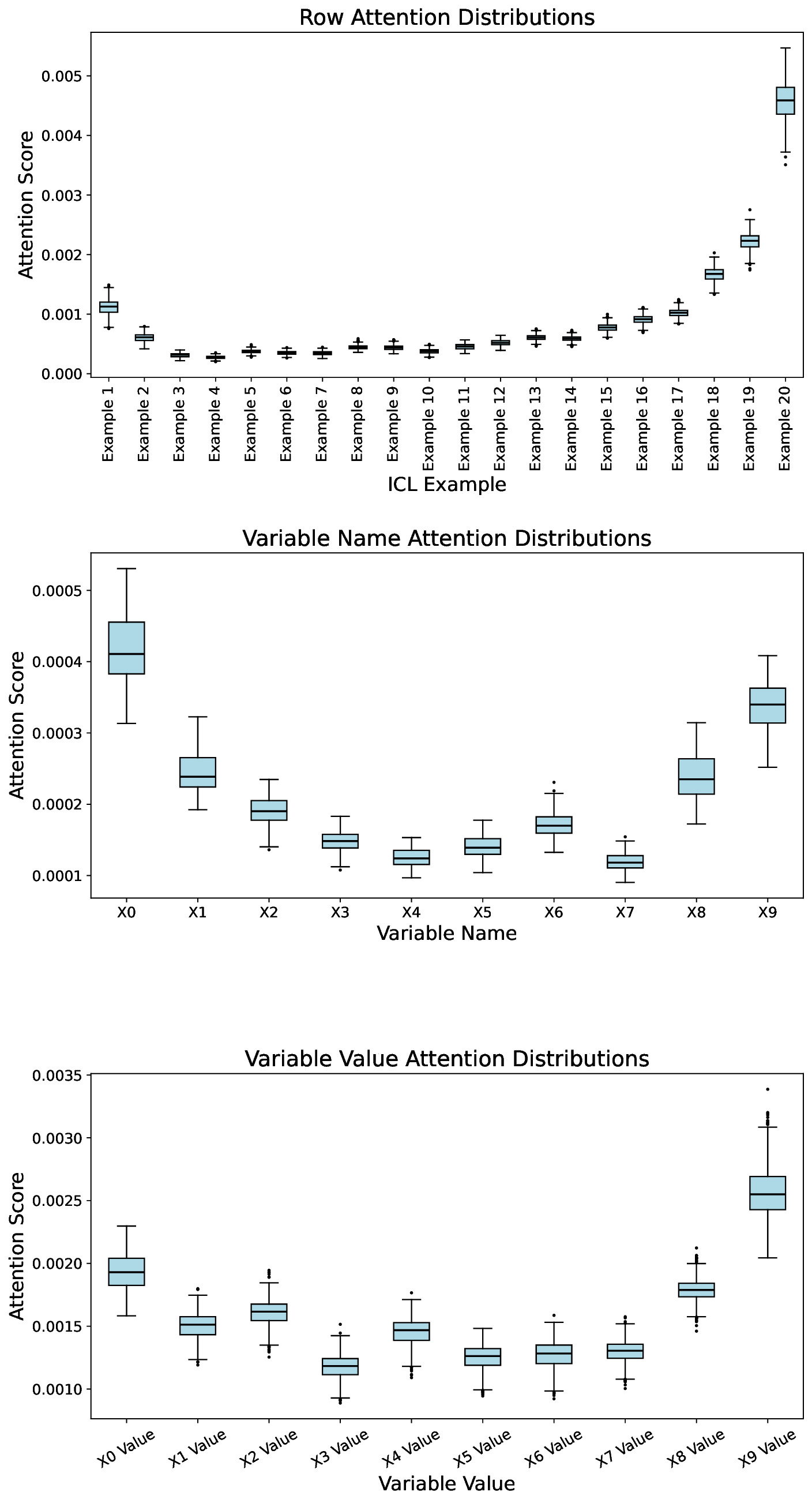}
        \caption{Attention Distributions by Rows, Variable Names, and Variable Values (Llama-3-8B-instruct under ICL20 setting)}
        \label{fig:attention_icl20}
    \end{minipage}%
    \hfill
    \begin{minipage}{0.49\textwidth}
        \centering
        \includegraphics[width=\linewidth]{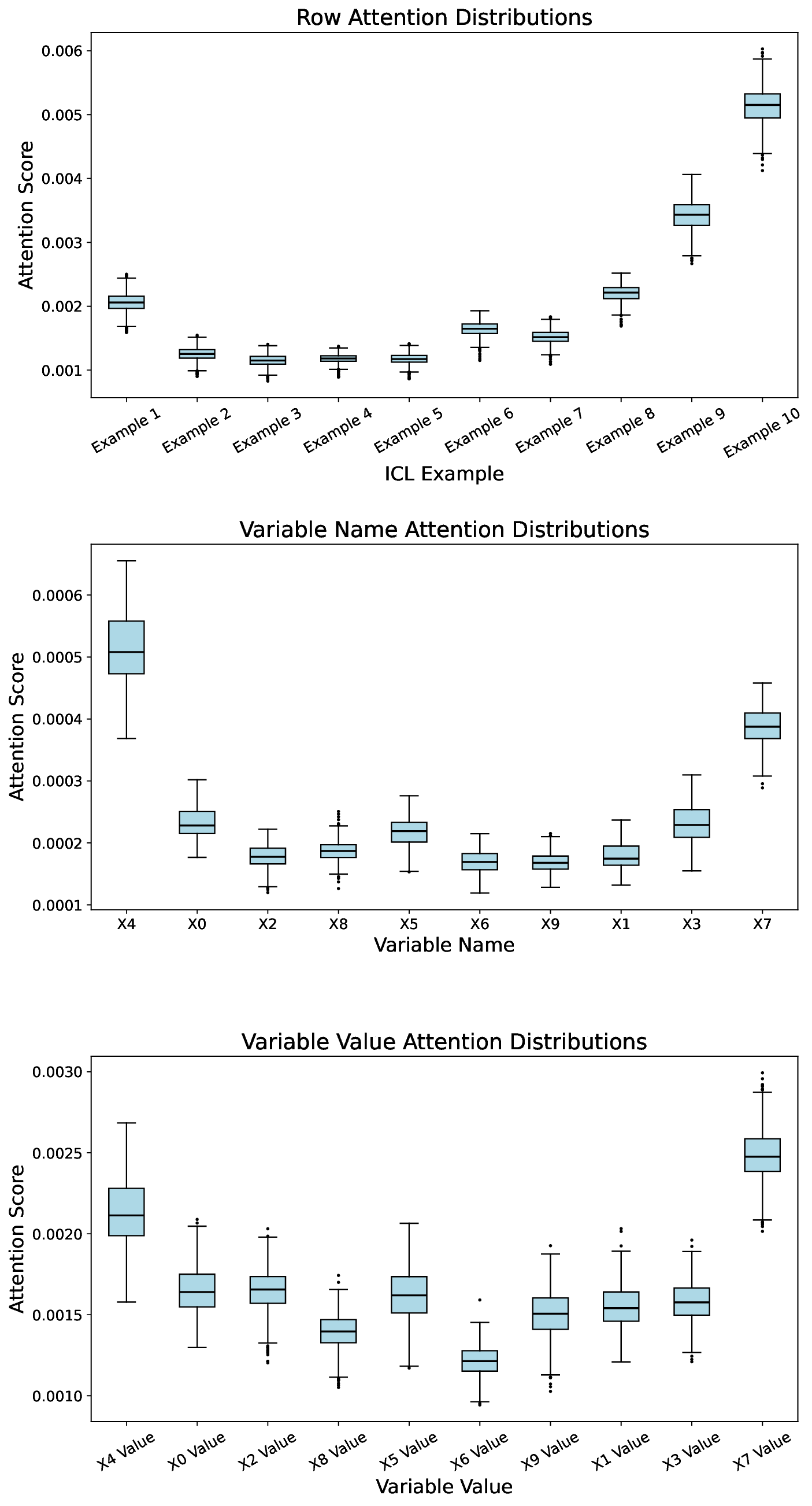}
        \caption{Attention Distributions by Rows, Variable Names, and Variable Values (Llama-3-8B-instruct under ICL10 setting with randomly shuffled variable order)}
        \label{fig:attention_icl10_shuffle}
    \end{minipage}
\end{figure}

Figure \ref{fig:attention_icl20} presents the attention distributions associated with the ICL20 setting. Figure \ref{fig:attention_icl10_shuffle} presents the attention distributions associated with the ICL10 setting with randomly shuffled variable orders. The variable order after the random shuffling is `X4', `X0', `X2', `X8', `X5', `X6', `X9', `X1', `X3', `X7'.

\end{document}